\documentclass[lettersize,journal]{IEEEtran}
\usepackage{amsmath,amsfonts}
\usepackage{algorithmic}
\usepackage{algorithm}
\usepackage{array}
\usepackage[caption=false,font=normalsize,labelfont=sf,textfont=sf]{subfig}
\usepackage{textcomp}
\usepackage{stfloats}
\usepackage{booktabs}
\usepackage{url}
\usepackage{enumitem}
\usepackage{verbatim}
\usepackage{float}
\usepackage{hyperref}
\usepackage{url}
\usepackage{graphicx}
\usepackage{subcaption}
\hypersetup{breaklinks=true}
\usepackage{graphicx}
\usepackage{enumitem}
\usepackage{amsmath} 

\usepackage[style=ieee,sorting=none]{biblatex}
\begin{document}
\title{FloorSAM: SAM-Guided Floorplan Reconstruction with Semantic-Geometric Fusion}
\author{Han Ye, Haofu Wang, Yunchi Zhang, Jiangjian Xiao, Yuqiang Jin, Jinyuan Liu, Wen-An Zhang, Uladzislau Sychou, Alexander Tuzikov, Vladislav Sobolevskii, Valerii Zakharov, Boris Sokolov, Minglei Fu
 \thanks{This work was supported by the National Key Research and Development Program of China under Grant No. 2022YFE0121700 and by the Russian Science Foundation under Grant No. FFZF-2025-0020 for V. Sobolevskii, V. Zakharov, and B. Sokolov. (\textit{Corresponding author: Minglei Fu}).}
 \thanks{H. Ye, Hanfu Wang,Yunchi Zhang, Y. Jin, J. Liu, Wen-An Zhang and M. Fu are with College of Information Engineering, Zhejiang University of Technology, Hangzhou 310023, China, and with the Joint Laboratory of Industrial Intelligent Technologies, Russia, China (e-mail: \{hye, haofuwang, yunchizhang, yqjin, jy.liu,wazhang, fuml\}@zjut.edu.cn).}
 \thanks{Jiangjian Xiao is with the Ningbo Institute of Materials Technology and Engineering, Chinese Academy of Sciences, Ningbo 315201, China (e-mail: xiaojj@nimte.ac.cn).}
 \thanks{Uladzislau Sychou, Alexander Tuzikov are with the United Institute of Informatics Problems of the National Academy of Belarus, Minsk 220012, Belarus (e-mail: vsychyov@robotics.by, tuzikov@newman.bas-net.by).} 
 \thanks{V. Sobolevskii, V. Zakharov, and B. Sokolov are with the Saint Petersburg Federal Research Center of the Russian Academy of Sciences, St. Petersburg, Russia, and with the Joint Laboratory of Industrial Intelligent Technologies, Russia, China (e-mail:  arguzd@yandex.ru, valeriov@yandex.ru, sokolov\_boris@inbox.ru).}
}
\markboth{}%
{Shell \MakeLowercase{\textit{et al.}}: A Sample Article Using IEEEtran.cls for IEEE Journals}
\maketitle

\begin{abstract}
Reconstructing building floor plans from point cloud data is a critical technology for indoor navigation, building information modeling (BIM), and highly accurate precise indoor measurement applications. Traditional methods, such as geometric algorithms and Mask R-CNN-based deep learning for mask segmentation, often suffer from sensitivity to noise, limited generalization, and loss of geometric details, severely impacting measurement accuracy. This study proposes an innovative framework, FloorSAM, that integrates room-height point cloud density maps with the guided segmentation capabilities of the Segment Anything Model (SAM) to enhance the precision of floor plan reconstruction from LiDAR point cloud data. By applying grid-based filtering to retain elevation point clouds near the ceiling of each region, combined with adaptive resolution projection and image enhancement techniques, a top-down density map is generated, improving the robustness and accuracy of spatial feature measurement. This framework leverages SAM’s zero-shot learning to achieve high-fidelity room segmentation, remarkably enhancing reconstruction and measurement accuracy across diverse building layouts. Subsequently, leveraging SAM’s zero-shot guided segmentation capabilities, high-quality room masks are generated based on adaptive prompt points, followed by a multistage filtering process to extract precise semantic masks for individual rooms. Through joint analysis of mask and point cloud modalities, contour extraction and regularization are performed, integrating semantic segmentation with geometric information to produce accurate room floor plans and recover topological relationships between rooms. Experiments on the Giblayout and ISPRS public datasets validate the effectiveness of our method, showing significant improvements in measurement accuracy, recall, and robustness over traditional approaches—especially in noisy environments and complex layouts. Our code, videos, and supplementary materials are available at https://github.com/Silentbarber/FloorSAM.

\end{abstract}
\begin{IEEEkeywords}
Floorplan Reconstruction, Semantic Segmentation, Zero-Shot Learning, Indoor Measurement.
\end{IEEEkeywords}

\setlength{\textfloatsep}{5pt}
\setlength{\floatsep}{5pt}
\maketitle

\begin{figure}[H]
    \centering
    \includegraphics[width=0.85\linewidth]{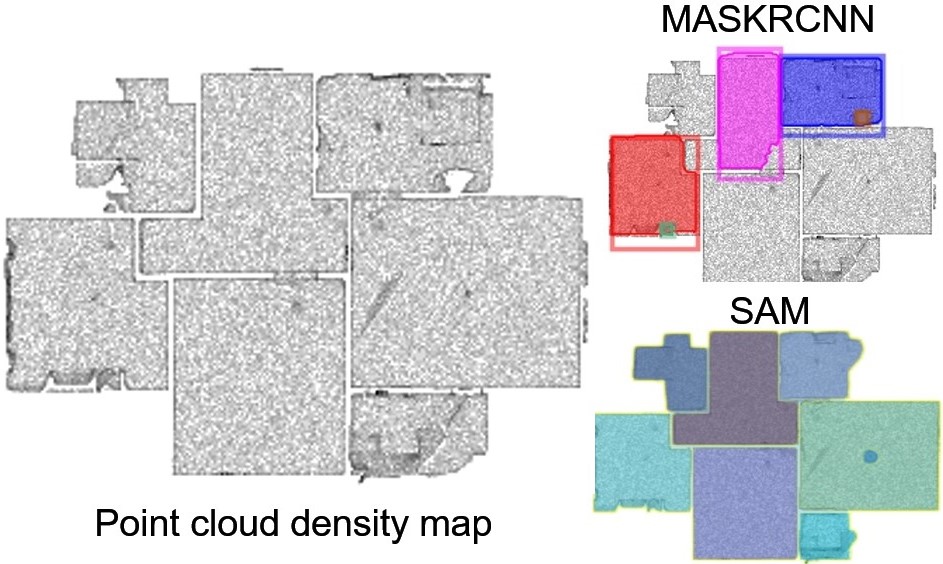}
    \caption{Comparison of the segmentation results produced by Mask R-CNN and SAM.}
    \label{sam+maskrcnn}
\end{figure}

\section{Introduction}
\IEEEPARstart{I}{n} recent years, the rapid advancement of 3D scanning technology, particularly LiDAR, has made point cloud data a vital resource for indoor scene reconstruction and precise measurement, owing to its high resolution and rich geometric detail \cite{10466622}. Accurate indoor scene reconstruction is critically important for applications such as BIM, indoor navigation, and safety assessment \cite{10598200,10132570}.However, automatically generating indoor floor plans from point cloud data remains challenging due to factors such as noise interference, occlusions, loss of geometric detail, and insufficient semantic information\cite{hu2023robot}. Current methods exhibit significant limitations in geometric robustness, data dependency, and adaptability to diverse scenes, highlighting the urgent need for innovative solutions.
\par Traditional geometric methods rely on hand-crafted geometric constraints (e.g., plane fitting, line detection) to reconstruct floorplans by detecting geometric primitives such as planes and lines \cite{cui2019automatic,wang2018semantic}. However, these methods demonstrate poor robustness in noisy and occluded scenes and struggle to recover semantic information, suffering from fundamental limitations. First, they heavily depend on the Manhattan assumption (e.g., orthogonal walls, horizontal ceilings), which makes it challenging to handle irregular structures such as curved walls and slanted roofs. For example, RANSAC-based line segmentation methods often misidentify non-wall lines or merge discontinuous segments in polyline wall scenarios \cite{gao2024floor}. Second, these methods are highly sensitive to noise and occlusions, where discrete points in LiDAR scans or furniture occlusions often lead to contour fragmentation. For instance, the Alpha Shape algorithm frequently fails to preserve true boundaries in regions with voids \cite{edelsbrunner2003shape}.

\par Deep learning-based methods have significantly advanced semantic segmentation. However, substantial limitations remain. Instance segmentation approaches, such as Mask R-CNN \cite{he2017mask}, can generate semantic masks but require large amounts of annotated training data and exhibit limited generalization capabilities. Studies have shown that Mask R-CNN-based methods perform well on specific datasets but experience considerable performance degradation in novel scenes, with accuracy dropping by over 30\% in unseen layouts (e.g., novel room structures)  \cite{jin2023reconstructing}. We trained Mask R-CNN on the 3D-Front dataset \cite{fu20213d} and compared its performance against the SAM \cite{kirillov2023segment} on a sample from the Giblayout dataset, as illustrated in Figure~\ref{sam+maskrcnn}. The results indicate that Mask R-CNN trained on 3D-Front data performs noticeably worse than SAM. Additionally, post-processing techniques such as GAN-based inpainting, commonly used to repair missing regions, often compromise geometric details, resulting in overly regularized reconstructions that deviate from the actual scene and lose critical features like non-orthogonal wall corners and door/window recesses \cite{jin2023reconstructing}.
\par To address these challenges and enhance measurement reliability, this study proposes an innovative framework, FloorSAM, that integrates room-height point cloud density maps with the guided segmentation capabilities of the Segment Anything Model (SAM) to enhance the precision of floor plan reconstruction from LiDAR point cloud data. The proposed method offers several notable advantages. Through grid-based filtering to retain high-elevation points near the ceiling, combined with adaptive resolution projection and image enhancement techniques, we generate bird's-eye view density maps. These maps significantly enhance spatial feature representation and improve robustness to noise. Leveraging SAM's zero-shot segmentation capability, high-quality room masks are generated based on adaptive prompt points, effectively eliminating the dependence on annotated data typical of traditional methods while providing strong generalization performance. Furthermore, by jointly processing the masks and point clouds during the contour extraction stage, the method integrates semantic segmentation results with geometric information, preserving fine details while regularizing boundaries. This enables accurate room floor plan generation and recovery of the topological relationships between rooms.
\par The main contributions of this work are summarized as follows:
\begin{enumerate}[leftmargin=2em, itemindent=0em, labelindent=0em] 
    \item We propose a hybrid framework that combines density maps and SAM, introducing zero-shot segmentation into point cloud floor plan reconstruction. This approach effectively addresses the reliance on annotated data inherent in traditional measurement methods and enhances measurement adaptability.
    \item We design a multi-stage mask filtering mechanism that extracts high-quality single-room segmentation results from the redundant masks generated by SAM, utilizing geometric constraints and semantic validation to improve measurement precision.
    \item We develop a mask–point cloud joint contour drawing and regularization algorithm that preserves geometric details while generating regularized contours, providing a reliable foundation for recovering topological relationships and ensuring measurement accuracy.
    \item Extensive experimental results demonstrate that FloorSAM outperforms state-of-the-art methods on various building point clouds. We have released the code for our method to facilitate further research in this field.
\end{enumerate}
\begin{figure*}
    \centering
    \includegraphics[width=0.9\textwidth]{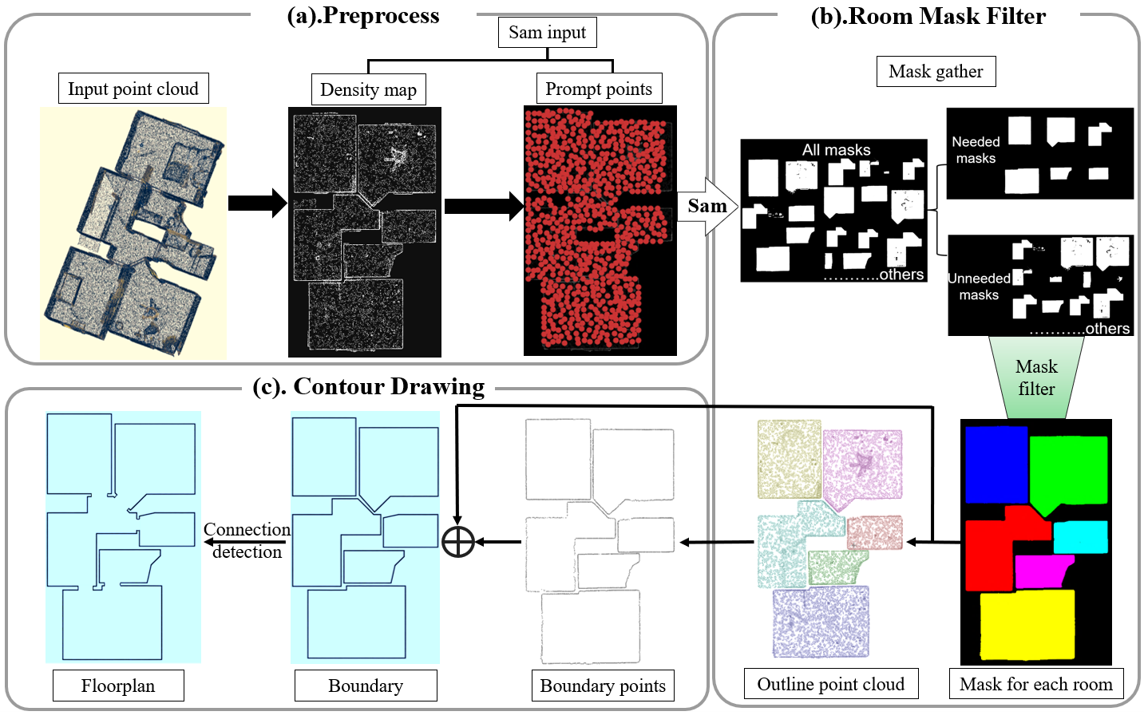}
    \caption{The overall architecture of FloorSAM. The first step is preprocessing (a), where a prompt map is extracted from ceiling density points for SAM segmentation. The second step is room mask filtering (b), where the segmentation results are refined through mask screening and filtering. The third step is contour drawing (c), where joint drawing by fusing semantic and geometric information regularizes contour and topology recovery.}
    \label{overview}
\end{figure*}
\section{RELATED WORKS}
In this section, we summarize classical methods for reconstructing floorplans from indoor point clouds. These methods can be broadly categorized into two types: traditional approaches and deep learning-based techniques.
\subsubsection{Traditional Methods}
Traditional methods for floorplan reconstruction from point clouds or images primarily utilize low-level image processing, geometric analysis, and optimization techniques. However, they often lack robustness and adaptability in complex indoor environments. Early approaches focused on extracting geometric primitives using techniques such as Hough transforms\cite{9469927} to detect linear features or superpixel segmentation \cite{qin2018accurate} to identify coherent regions, enabling the reconstruction of wireframes \cite{11005571} or polygonal loops \cite{gimenez2015reconstruction} from rasterized data. These methods heavily rely on handcrafted features and heuristic thresholds, resulting in poor generalization across diverse scenes. 
\par More advanced techniques employ optimization-based frameworks, such as graph-cut-based graphical model inference \cite{silberman2012indoor}, dynamic programming \cite{chen2019floor}, or Bayesian networks \cite{delage2006dynamic}, to infer room layouts and topological structures. While these approaches improve structural coherence, their dependence on manually tuned parameters limits robustness, especially in scenes with irregular geometries.
\par Surface-based methods, which utilize algorithms such as RANSAC \cite{schnabel2007efficient} or region growing \cite{rabbani2006segmentation} for planar element extraction, aim to identify structural components like walls and openings \cite{9721911}. However, these methods are highly sensitive to noise and occlusions, often failing to capture accurate boundaries in cluttered environments. Spatial partitioning techniques, such as point cloud projection followed by morphological operations \cite{tang2022bim} or ray-casting combined with visibility analysis \cite{chen2024semantic}, enhance robustness by generating binary images or segmented spaces. Despite better handling of complex layouts, these methods may introduce boundary inaccuracies or fail to preserve the topological relationships between rooms. 
\par Graph structure inference methods adopt a bottom-up approach by detecting corners and edges \cite{liu2017raster, cabral2014piecewise} and employing dynamic programming or integer programming \cite{lee2017roomnet, schwing2012efficient} to reconstruct topological structures. Nevertheless, these techniques are prone to detection errors in complex scenes, resulting in incomplete or incorrect floorplan reconstructions. Collectively, traditional methods struggle with noise, occlusions, and non-Manhattan structures, underscoring the need for more robust and adaptive approaches that effectively integrate semantic and geometric information.

\subsubsection{Deep Learning-Based Methods}
In recent years, deep learning techniques have significantly enhanced the robustness and efficiency of floor plan reconstruction and are widely employed to infer geometric and topological structures from point cloud data\cite{9153026,10174665}. Early approaches, such as those by Nauata et al. \cite{nauata2020vectorizing} and Floor-SP \cite{chen2019floor}, utilize a two-stage pipeline: first, Mask R-CNN detects corners, edges, or regions; then, integer programming assembles the floor plan. While these methods yield good results, they suffer from slow inference speeds and are sensitive to noise and occlusion, often resulting in missing walls or rooms in complex scenes due to detection failures.
\par HEAT \cite{chen2022heat} proposed an attention-based Transformer architecture that processes 2D raster images to reconstruct the underlying geometric structure. It demonstrates strong noise resistance but has limited generalization to irregular room layouts. RoomFormer \cite{yue2023connecting} introduced a novel Transformer architecture featuring a two-level query mechanism (room-level and corner-level) to generate polygon sequences in parallel. This approach avoids the Manhattan-world assumption, supports end-to-end training, and achieves state-of-the-art performance on the Structured3D and SceneCAD datasets with fast inference speed. However, it still faces challenges when handling heavily occluded scenes.\par PolyGraph \cite{sun2025polygraph} begins with 2D density or normal maps generated from 3D point clouds and detects wall points using a cross-guided neural network. It optimizes subgraphs to reduce the search space and improve reconstruction quality. Its wall point primitives are more tolerant of errors than traditional corner-based methods; however, uneven wall point distribution in complex scenes can negatively impact accuracy.PolyRoom \cite{liu2024polyroom} introduces a room-aware transformer approach that enhances reconstruction accuracy in complex scenes through uniform sampling representation and self-attention mechanisms. While it demonstrates strong generalization capabilities, it still struggles to detect very small rooms.\par FRI-Net \cite{xu2024fri} introduced a room-level implicit representation method that combines geometric priors with a two-stage training strategy—initially optimizing horizontal and vertical lines, followed by incorporating diagonals to capture angular features—to generate coherent and closed polygons with strong robustness. However, this approach may overlook high-frequency local details in internal wall reconstruction due to the limitations of implicit representation. The application of deep learning in floor plan reconstruction still requires further advancements in generalization and detail preservation, especially under complex scenarios.

\section{METHODOLOGY}
\subsection{Overview}
As illustrated in Figure~\ref{overview}, the proposed FloorSAM method converts point clouds into floorplans through three steps: First, a density map is generated from point clouds near the ceiling after simple denoising. Then, numerous prompt points are created based on the ceiling map. The density map and prompt points serve as inputs to SAM for comprehensive segmentation. Next, a coarse-to-fine two-step filtering process is applied to select high-quality single-room masks that represent the semantics of each room. Finally, room semantics and their corresponding projected point clouds are jointly used to draw regularized contours and detect connection segments for topological recovery.

\subsection{Density Map and SAM Integration}

To achieve high-quality semantic mask segmentation of single rooms, traditional methods typically rely on geometric fitting of roofs or walls or use manually annotated data to train instance segmentation models such as Mask R-CNN. However, regular wall surfaces in building point clouds are often affected by noise, and Mask R-CNN’s generalization capabilities are limited, making it difficult to adapt to the diversity of complex indoor scenes. 
\par To address this challenge, we propose a segmentation framework that combines point cloud density maps with the SAM. By generating bird's-eye view (BEV) density maps and leveraging zero-shot guided segmentation, our method efficiently extracts high-quality semantic masks for indoor scenes. We employ an adaptive prompt point generation strategy to ensure SAM's segmentation results encompass all potential high-quality single-room masks, making it particularly suitable for complex room layouts with non-Manhattan structures \cite{xu2023sampro3d}. Additionally, by retaining elevation point clouds near the ceiling, the spatial feature representation of the density map is enhanced, thereby improving segmentation robustness. 
\begin{figure*}
    \centering
    \includegraphics[width=0.85\textwidth]{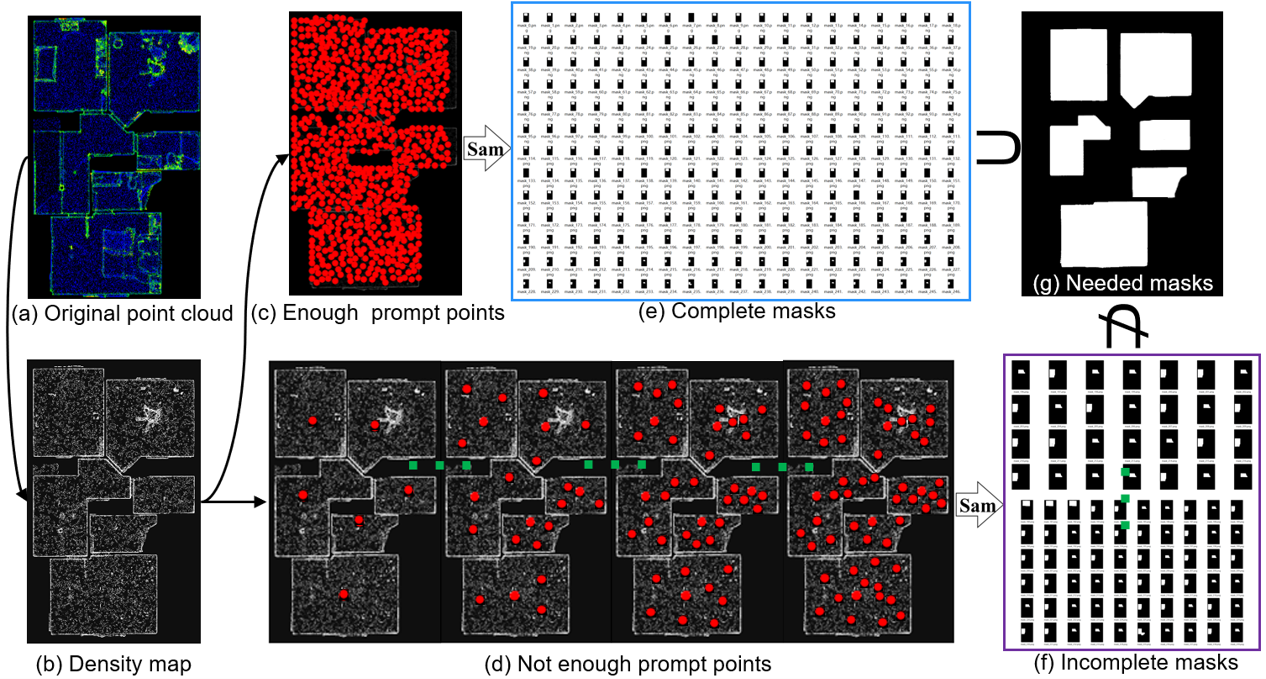}
    \caption{The impact of different prompt points on segmentation. (a) is the original input point cloud; (b) is the adaptive density point cloud map; (c) is a sufficient amount of SAM adaptive prompt points; (d) is a variety of insufficient SAM prompt point samples; (e) is a mask containing the correct mask segmented by SAM using sufficient adaptive prompt points, which completely contains the mask in (g); (f) is an incomplete mask segmented by SAM using insufficient amount of prompt points, which does not completely contain the mask in (g); (g) is the mask of each room required for subsequent floor plan drawing.}
    \label{amethod}
\end{figure*}

\subsubsection{Point Cloud Preprocessing and Grid-Based Filtering}

To reduce the effects of noise and occlusions, the point cloud data is preprocessed to retain elevation points near the ceiling. Let the point cloud be \( P = \{ p_i = (x_i, y_i, z_i) \mid i=1, \dots, N \} \). A grid-based method divides the XOY plane into grids with a grid size \(\gamma = 0.1 \, \text{m}\). Within each grid \((m, n)\), points with the highest \(z\)-value are retained, satisfying a height tolerance \(\Delta z \leq 0.1 \, \text{m}\), to filter out low objects (e.g., furniture). The filtered point cloud is defined as:
\begin{equation}
P' = \{ p_j' \mid z_j \geq \max(z) - 0.1, p_j' \in \text{grid}(m, n) \}.
\end{equation}
This step effectively reduces point cloud complexity while preserving points near walls, enhancing the spatial feature representation for subsequent density map generation.

\subsubsection{Adaptive Resolution Density Map Generation}

The filtered point cloud \( P' \) is projected onto the XOY plane to generate a 2D bird's-eye view (BEV) density map, as illustrated in Figure~\ref{amethod}(b). Let the point cloud bounding range be \([x_{\min}, x_{\max}] \times [y_{\min}, y_{\max}]\). The target resolution \( R \) is determined using a resolution factor \(\kappa\) (default value 1000) and the point cloud's average point spacing \(\delta\):
\begin{equation}
R = \kappa \cdot \frac{\max(x_{\max} - x_{\min}, y_{\max} - y_{\min})}{\delta}.
\end{equation}
The scale factor \( s \) is computed based on \( R \) to ensure compatibility with SAM's input requirements. The density map \( D(m, n) \) is constructed by counting the density of projected points:

\begin{equation}
\begin{split}
D(m, n) &= \sum_{p_j' \in P'} \mathbb{1} \Big( \lfloor \frac{x_j - x_{\min}}{s} \rfloor = m, \\
&\quad \lfloor \frac{y_j - y_{\min}}{s} \rfloor = n \Big),
\end{split}
\end{equation}
where \(\mathbb{1}(\cdot)\) is the indicator function. The density map is then processed with logarithmic transformation \( D'(m, n) = \log(1 + D(m, n)) \), Gaussian blur (kernel size \(\sigma = 5\)), and CLAHE enhancement to improve robustness and contrast of spatial features, providing high-quality input for prompt point extraction and segmentation.

\subsubsection{Adaptive Prompt Point Extraction}

To ensure SAM segmentation covers all potential single-room regions, an adaptive prompt point generation strategy is designed, as illustrated in Figure~\ref{amethod}(c). Gaussian blur and max-pooling are applied to the enhanced density map \( D' \) to detect local density peaks, selecting high-density regions as candidate prompt points. The density peak set is defined as:
\begin{equation}
Q = \{(m_k, n_k) \mid D'(m_k, n_k) \geq \tau \},
\end{equation}
where the threshold \(\tau = 0.9 \cdot \text{mean}(D')\). The prompt point set is optimized by enforcing a minimum distance constraint of 10 pixels to ensure uniform distribution, covering all potential room regions and providing sufficient guidance for SAM.

We illustrate the effect of our adaptive prompt point generation in Figure~\ref{amethod}. In Figure~\ref{amethod}(e), the adaptive prompt points produce dense and accurate room-wise semantic masks, many of which precisely delineate complex room boundaries. In contrast, Figure~\ref{amethod}(d) only assigns a small number of prompt points to each room, resulting in incomplete segmentation and low accuracy, and the final result does not fully contain all the required room semantic masks. This comparison underscores the effectiveness and robustness of our adaptive strategy, particularly in handling complex room structures.

\subsubsection{SAM-Guided Segmentation}

Leveraging SAM’s zero-shot segmentation capability, the enhanced density map \( D' \) is used as input, with the prompt point set generating multiscale masks. SAM produces multiple candidate masks for each prompt point, enhancing segmentation diversity through a multimask output strategy to ensure high-quality single-room masks, as illustrated in Figure~\ref{amethod}(d) (f). The segmentation results are saved as independent mask images for subsequent filtering and processing. This step exploits SAM's generalization ability, eliminating the need for annotated data and adapting to complex and diverse indoor scenes.

\subsection{High-Quality Single-Room Mask Filtering}

\begin{figure}
    \centering
    \includegraphics[width=1\linewidth]{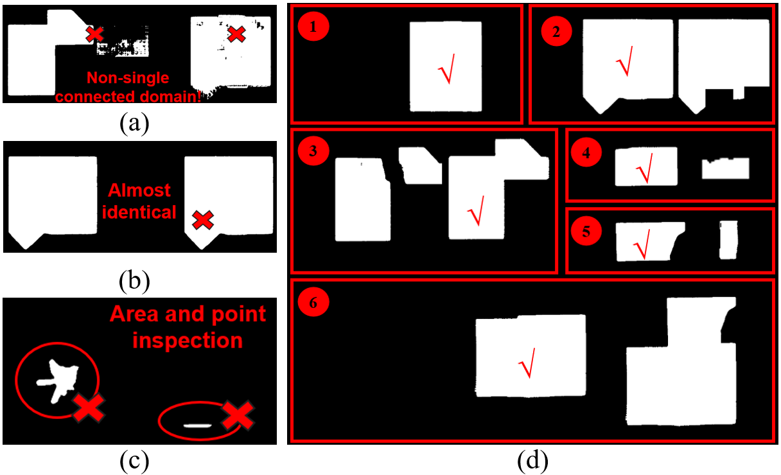}
    \caption{Coarse and fine filtering of semantic masks. (a) Removal of masks with non-single connected domains; (b) elimination of highly similar masks via IoU-based deduplication; (c) filtering of masks with inadequate area or point cloud counts; (d) grouping and refinement for fine mask selection.}
    \label{bmethod}
\end{figure}

To extract high-quality single-room semantic masks from the redundant masks generated by SAM, we propose a multi-stage filtering mechanism that integrates geometric constraints, semantic validation, and combinatorial optimization. This approach effectively identifies masks that accurately represent individual rooms, ensuring robustness in complex indoor scenes with non-Manhattan structures. The filtering process comprises coarse and fine stages, leveraging point cloud geometry and mask semantics to produce a refined set of masks suitable for subsequent contour extraction and topological relationship recovery, as illustrated in Figure~\ref{bmethod}.

\subsubsection{Coarse Filtering Stage}

The coarse filtering stage rapidly eliminates masks that deviate from single-room characteristics by applying geometric and point cloud constraints. Masks are first evaluated for connectivity, those with multiple components or excessive holes (e.g., more than two) are discarded, as shown in Figure~\ref{bmethod}(a). Redundant masks are then removed using an Intersection over Union (IoU) criterion, where masks with IoU \(\geq 0.8\) are compared, retaining only the mask with the larger area or higher point cloud point count, as depicted in Figure~\ref{bmethod}(b). Additionally, masks are assessed based on their area and corresponding point cloud points. Masks with areas below 2\% of the total scene area or 1 square meter, or exceeding a dynamic threshold based on the 75th percentile plus twice the interquartile range, or with insufficient point counts, are eliminated, as illustrated in Figure~\ref{bmethod}(c). This stage efficiently reduces the candidate set by removing clearly erroneous or redundant masks.

\subsubsection{Fine Filtering Stage}

To address remaining problematic masks, such as those representing incomplete rooms or combining multiple rooms, the fine filtering stage employs a sophisticated strategy involving grouping, inclusion analysis, and combinatorial optimization. Masks are grouped by constructing an adjacency matrix \( G \), where \( G_{ij} = 1 \) if \(\text{IoU}(M_i, M_j) \geq 0.5\), and connected component analysis identifies semantically similar mask sets, as shown in Figure~\ref{bmethod}(d). For instance, six groups are identified, with the second group containing a redundant incomplete room mask, the third through fifth groups including partial room masks, and the sixth group featuring a composite mask merging two rooms. Inclusion analysis detects and removes redundant or composite masks; if \( |M_i \cap M_j| / |M_j| \geq 0.9 \), mask \( M_j \) is deemed included in \( M_i \) and discarded, alongside composite masks with abnormal areas or multiple sub-rooms. Within each group, the mask with the largest area or the one encompassing others is selected to ensure semantic completeness. Finally, a greedy algorithm optimizes the mask combination to maximize the number of masks while covering at least 95\% of the total candidate area:
\begin{equation}
\sum_{M_i \in S} A_i \geq 0.95 \cdot A_{\text{total}},
\end{equation}
with an IoU tolerance of 0.01 (\(\text{IoU}(M_i, M_j) \leq 0.01\)) to ensure minimal overlap, yielding a high-quality single-room mask set \( S \) that comprehensively covers the scene.

This multi-stage filtering mechanism balances geometric rigor and semantic integrity, demonstrating robust performance in complex layouts and noisy environments, as evidenced by the refined mask sets in Figure~\ref{bmethod}, which provide a solid foundation for subsequent contour extraction and floor plan generation.

\subsection{Mask-Point Cloud Joint Contour Extraction and Regularization}

After obtaining semantic masks for each room, we propose a mask-point cloud joint contour extraction and regularization algorithm to accurately extract room contours and generate regularized floor plans from point cloud data and SAM-generated semantic masks, as illustrated in Figure~\ref{cmethod1}. By integrating mask semantic information with the geometric properties of the point cloud, high-quality vector contours are produced, facilitating the recovery of inter-room topological relationships. To address potential mask contour deviations and over-regularization issues, point cloud boundary points are used for correction, ensuring both geometric accuracy and semantic consistency. The detailed process is as follows:

\begin{figure}
    \centering
    \includegraphics[width=0.8\linewidth]{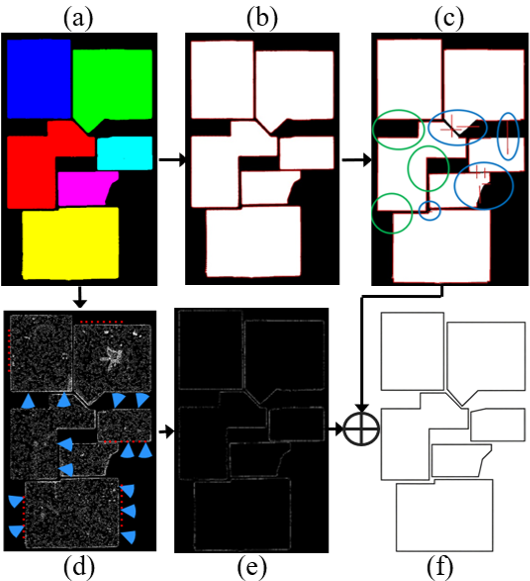}
    \caption{The overall process of mask-point cloud joint contour extraction and regularization. (a) is the correct mask of each room screened out in the previous step; (b) is the detected contour after RDP simplification; (c) is the regularized contour, where the blue circle is the wrong regularized area, and the green circle is the correct regularized area; (d) is the room point cloud corresponding to the semantics of each room, and the blue sector in the figure is the detection area of each room boundary point; (e) is the boundary point extracted from each room; (f) is the final regularized contour drawn jointly by (c) and (e).}
    \label{cmethod1}
\end{figure}

\begin{figure}
    \centering
    \includegraphics[width=1\linewidth]{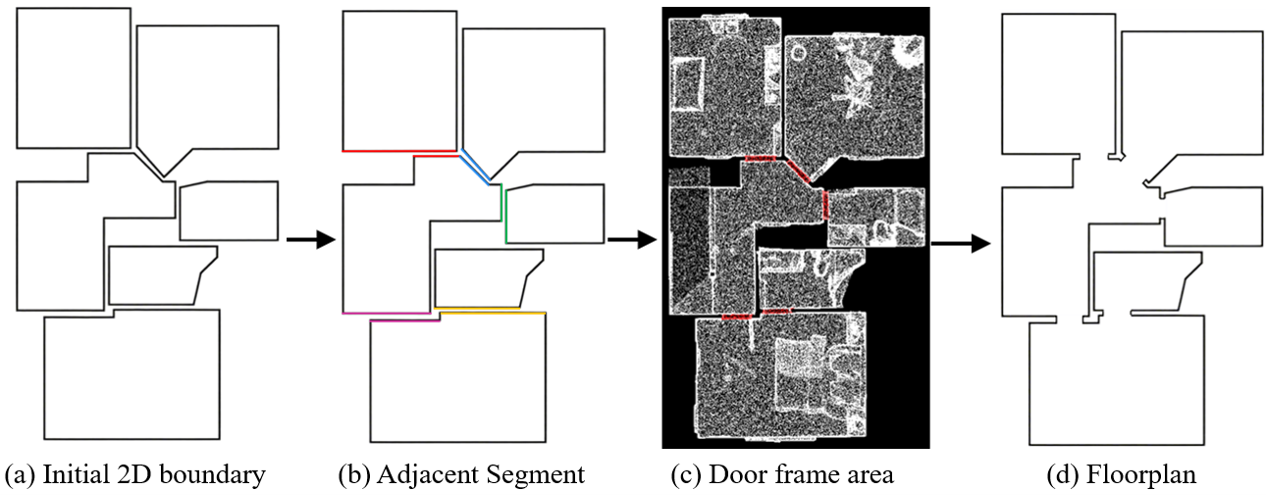}
    \caption{The process of restoring connectivity between houses. (a) initial regularized outline; (b) detected adjacent line segments of each room; (c) door frame point cloud between adjacent line segments; (d) final floor plan after topological relationship restoration.}
    \label{cmethod4}
\end{figure}

\subsubsection{Point Cloud Projection and Boundary Extraction}

The point cloud \( P \) is projected onto the XOY plane (\( z=0 \)), and the ceiling plane region is extracted using RANSAC, generating a 2D projected point cloud \( P' = \{ (x_i, y_i) \mid i=1, \dots, N' \} \). KD-Tree neighborhood analysis and angular resolution checks identify boundary points, as shown in Figure~\ref{cmethod1}. A point \( p_i \)'s neighborhood is defined as \( N(p_i, r) \). If a continuous angular sector (e.g., 30 degrees) within radius \( r \) (e.g., 0.2 m) contains no points, \( p_i \) is marked as a boundary point, forming the boundary point set \( B_p = \{ b_j \mid j=1, \dots, N_b \} \).

\subsubsection{Mask Contour Extraction and Initial Processing}

From the filtered single-room masks, OpenCV’s contour detection algorithm extracts boundary points, forming an initial 2D contour. Let the boundary point set of mask \( M_i \) be \( B_i = \{ (x_j, y_j) \mid j=1, \dots, K \} \). The Ramer-Douglas-Peucker (RDP) algorithm simplifies the contour with a dynamic distance threshold \( \epsilon \) (adaptively adjusted based on point cloud density and mask area) to retain key geometric features and remove redundant points, generating a preliminary contour segment set \( L_i = \{ l_k \mid k=1, \dots, M \} \), as shown in Figure~\ref{cmethod1}(b).

\subsubsection{Joint Contour Generation and Correction}

The mask contour is combined with the point cloud boundary points to generate the final contour, with the point cloud boundary points correcting any deviations in the mask contour. Coordinates of mask contour points are assumed to be aligned with the point cloud through preprocessing.
\begin{itemize}[leftmargin=1em, labelsep=0.3em]
    \item \textbf{Main Direction Determination}: The main direction \( \theta_{\text{main}} \) is determined based on the longest contour segment or the segment with the smallest angle to the X/Y axis. Contour segments are classified as parallel (\( |\theta_k - \theta_{\text{main}}| < 45^\circ \)) or perpendicular (\( |\theta_k - \theta_{\text{main}}| \geq 45^\circ \)) to the main direction.
    \item \textbf{Direction Adjustment}: For each segment \( l_k \), compute its angle \( \theta_k \) with the main direction. If \( |\theta_k| < 45^\circ \), rotate clockwise or counterclockwise to \( \theta_{\text{main}} \); otherwise, rotate to \( \theta_{\text{main}} + 90^\circ \), preserving the original length to ensure directional consistency. Figure~\ref{cmethod1}(c) illustrates the direction adjustment, i.e., regularization effect with examples of both incorrect and correct outcomes. Relying solely on mask image contours for regularization produces suboptimal results, whereas using point cloud boundary points effectively guides contour simplification and regularization.
    \item \textbf{Contour Fusion and Correction}: Match mask contour points \( B_i' = \{ (x_j', y_j') \} \) with point cloud boundary points \( B_p \), using a loss function:
    \begin{equation}
    L(b_j') = \min_{b_p \in B_p} \| b_j' - b_p \|_2.
    \end{equation}
    If \( L(b_j') > \tau \) (threshold \( \tau \) typically 0.05 m), adjust \( b_j' \) to the nearest point cloud boundary point \( b_p \), ensuring contours adhere to geometric constraints while preserving mask semantic integrity. The complete effect of point cloud correction on image contours is shown in Figure~\ref{cmethod1}, demonstrating that the final contours are closer to reality.
\end{itemize}

\subsubsection{Room Topology Recovery}
As shown in Figure~\ref{cmethod4}, after obtaining regularized contours for each room, we use these contours to identify point clouds in the connection regions between adjacent rooms. The point clouds are then used to fit doorframe contour shapes, which are subsequently added to the rooms' regularized contours to recover the topological relationships between rooms.

\section{EXPERIMENT}
In this section, we introduce the two datasets used and present the experimental results of the proposed algorithm. The datasets include the GibLayout dataset\cite{wang2022building} and the ISPRS benchmark\cite{khoshelham2017isprs}. Additionally, we validated the algorithm's performance on self-collected data and showcased the results. The algorithm was tested on a total of 50 indoor scene datasets, and the results were compared with state-of-the-art methods. All experiments were conducted on a PC equipped with an 8-core AMD Ryzen 7 5800H CPU (3.2 GHz) and an NVIDIA GeForce RTX 3060 GPU.

\subsection{Datasets}
We evaluated the algorithm using two datasets: the GibLayout dataset and the ISPRS benchmark dataset. The GibLayout dataset contains 44 house models, each comprising 3 to 10 rooms, covering a total area of over 10,000 square meters. Its point cloud data is sourced from the public Gibson dataset\cite{xia2018gibson}, generated by NavVis using dual multi-layer LiDAR sensors and advanced SLAM algorithms, achieving map-level accuracy. We selected 12 scenes from the GibLayout dataset for testing, encompassing various roof types and indoor structures, including circular layouts, corridors, and non-Manhattan configurations. Additionally, we used the ISPRS Indoor Modeling Benchmark dataset, designed to advance point cloud-based 3D indoor modeling research, with point clouds collected by multiple sensors. We selected two representative scenes, TUB2 and UoM, for testing. Since our algorithm currently does not support multi-story building reconstruction, we extracted single-floor point clouds from both datasets for the experiments.

\subsection{Evaluation Metrics}
First, we evaluated the performance of room semantic segmentation. We recorded the number of correctly segmented room masks, denoted as \(room_{\text{true}}\), and the total number of room masks segmented by our method, denoted as \(room_{\text{all}}\). Using the ground truth 3D models from the GibLayout and ISPRS datasets, we manually annotated the corresponding ground truth room mask count, denoted as \(room_{\text{gt}}\). A room mask is considered correct if the overlap between the segmented room semantic mask and the manually annotated ground truth mask exceeds a threshold of 95\%.

Additionally, we used two metrics, precision and recall of contour edges in the reconstructed floorplan, to evaluate the floorplan drawing results. Using the ground truth 3D models from the GibLayout and ISPRS datasets, we manually drew the corresponding ground truth floorplans. As shown in Equations~\eqref{eq:precision} and \eqref{eq:recall}, \(boundary_{\text{true}}\) represents the number of correctly reconstructed line segments,  \(boundary_{\text{all}}\) denotes the total number of reconstructed line segments, and \(boundary_{\text{gt}}\) indicates the total number of line segments in the manually drawn ground truth floorplan. A line segment is considered correct if the distance between the endpoints of the reconstructed line and the corresponding ground truth line segment is below a threshold of 0.5 cm.
\begin{equation}
\label{eq:precision}
precision_{boundary} = \frac{boundary_{\text{true}}}{boundary_{\text{all}}}
\end{equation}
\begin{equation}
\label{eq:recall}
recall_{boundary} = \frac{boundary_{\text{true}}}{boundary_{\text{gt}}}
\end{equation}

\subsection{Dataset Experimental Results}

\subsubsection{GibLayout Dataset Results}

\begin{figure*}
\centering
\includegraphics[width=0.8\textwidth]{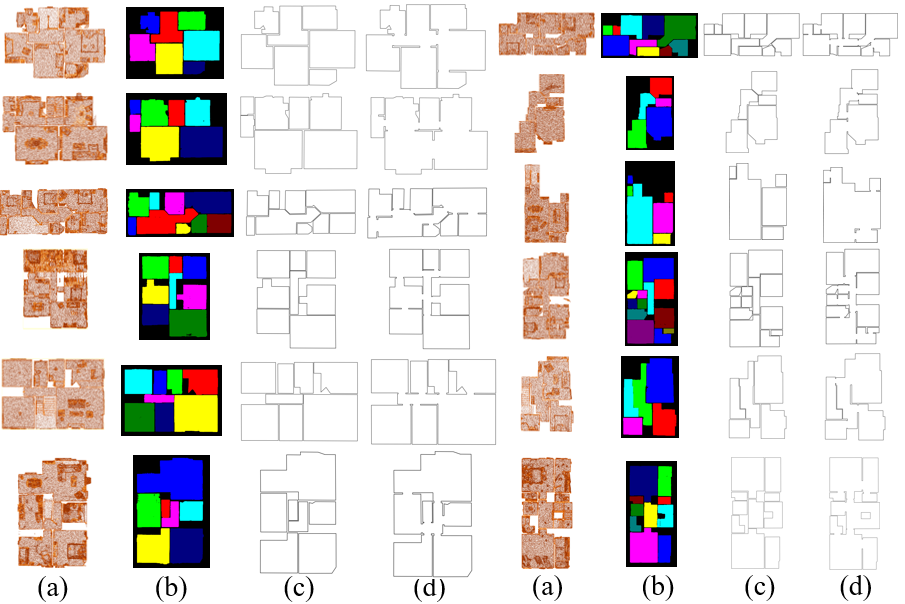}
\caption{Room segmentation and floorplan reconstruction results on the GibLayout dataset. (a) Input point cloud; (b) room segmentation results; (c) room contours; (d) floorplan with restored topological relationships.}
\label{experience1}
\end{figure*}

The experimental results of our algorithm on the GibLayout dataset are presented in Figure~\ref{experience1}. In the figure, (a) displays the input point cloud, (b) shows the segmented room semantic masks, (c) illustrates the reconstructed floorplan for each room with black lines representing walls, and (d) depicts the floorplan after recovering topological relationships. Our method effectively segments rooms, using distinct colors to differentiate individual room regions. For each room, we employed a mask-point cloud joint contour extraction technique, followed by the recovery of topological relationships between rooms through the detection of connection point clouds.
\par As shown in Table~\ref{table1}, our method successfully segmented nearly all rooms in the GibLayout dataset. However, in certain cases (e.g., scenes 3, 9, and 10), some masks were not segmented due to the absence of ceiling point clouds, leading to missing elements in the density map. Table~\ref{table2} summarizes the precision and recall of the floorplan reconstruction on the GibLayout dataset, achieving an average precision of 0.90 and an average recall of 0.94. While our approach effectively segments most room masks and reconstructs accurate room contours, performance in terms of precision and recall may be impacted when ceiling elevation point clouds are unavailable.

\begin{table}
  \centering
  \caption{Statistics of the number of rooms in the segmentation results on the GibLayout dataset.}
  \renewcommand{\arraystretch}{1.5} 
  \begin{tabular}{c|cccccccccccc}
    \hline
    Giblayout & 1 & 2 & 3 & 4 & 5 & 6 & 7 & 8 & 9 & 10 & 11 & 12 \\
    \hline
    $room_{true}$ & 6 & 7 & 9 & 8 & 8 & 7 & 10 & 5 & 5 & 12 & 5 & 10 \\
    $room_{all}$ & 6 & 7 & 9 & 8 & 8 & 7 & 10 & 5 & 6 & 13 & 5 & 10 \\
    $room_{gt}$ & 6 & 7 & 10 & 8 & 8 & 7 & 10 & 5 & 6 & 13 & 5 & 10 \\
    \hline
  \end{tabular}
  \label{table1}
\end{table}

\begin{table*}
  \centering
  \caption{The floor plan accuracy and recall of this method are tested on the GibLayout dataset.}
  \renewcommand{\arraystretch}{1.5} 
  \begin{tabular}{c|ccccccccccccc}
    \hline
    Giblayout & 1 & 2 & 3 & 4 & 5 & 6 & 7 & 8 & 9 & 10 & 11 & 12 & Mean \\
    \hline
    $Precision_{boundary}$ & 0.92 & 0.92 & 0.82 & 0.94 & 0.90 & 0.90 & 0.93 & 0.95 & 0.88 & 0.84 & 0.90 & 0.95 & 0.90 \\
    $Recall_{boundary}$ & 0.90 & 0.94 & 0.89 & 0.98 & 0.94 & 0.96 & 0.96 & 0.97 & 0.92 & 0.92 & 0.94 & 0.98 & 0.94 \\
    \hline
  \end{tabular}
  \label{table2}
\end{table*}
\subsubsection{ISPRS Dataset Results}
\begin{figure}
    \centering
    \includegraphics[width=1\linewidth]{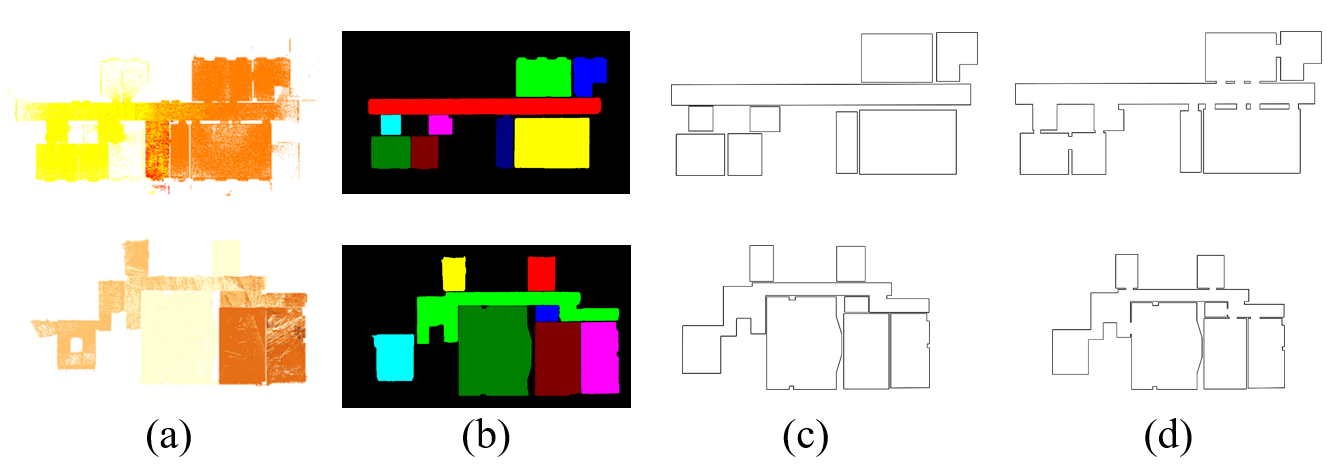}
    \caption{The room segmentation and floor plan reconstruction effects of the ISPRS dataset. (a) shows the input point cloud; (b) shows the result of room segmentation; (c) shows the outline of each room; (d) shows the topological relationship after restoration.}
    \label{experience4}
\end{figure}

The ISPRS benchmark dataset contains fewer scenes but larger ones compared to GibLayout, with an average area of several hundred square meters and corridors extending tens of meters. To improve computational efficiency, we applied voxel sampling to the point clouds of the two scenes, reducing the number of points using a voxel size of 0.05. The test results on this dataset are shown in Figure~\ref{experience4}. In the figure, (a) represents the input point cloud, (b) shows the segmented room semantic mask results, (c) displays the reconstructed floorplan for each room with black lines indicating walls, and (d) illustrates the floorplan after topological relationship recovery.  These results demonstrate that our method performs well on large-scale data, exhibiting strong generalization capabilities.

As shown in Table~\ref{table3}, our method accurately segmented all rooms in the ISPRS dataset. Table~\ref{table4} presents the floorplan results on the ISPRS benchmark dataset, demonstrating that our method maintains a precision above 0.90 and a recall above 0.96.
\begin{table}
  \centering
  \caption{Statistics of the number of rooms in the segmentation results on the ISPRS dataset.}
  \renewcommand{\arraystretch}{1.5} 
  \begin{tabular}{@{\hspace{3em}}c@{\hspace{3em}}|@{\hspace{3em}}c@{\hspace{3em}}c@{\hspace{3em}}}
    \hline
    ISPRS & TUB2 & UoM \\
    \hline
    $room_{true}$ & 9 & 8 \\
    $room_{all}$ & 9 & 8 \\
    $room_{gt}$ & 9 & 8 \\
    \hline
  \end{tabular}
  \label{table3}
\end{table}

\begin{table}
  \centering
  \caption{The precision and recall of the floor plan of this method are tested on the ISPRS dataset.}
  \renewcommand{\arraystretch}{2} 
  \begin{tabular}{@{\hspace{3em}}c|@{\hspace{3em}}c@{\hspace{3em}}c@{\hspace{3em}}c@{\hspace{3em}}}
    \hline
    ISPRS & TUB2 & UoM & Mean \\
    \hline
    $Precision_{boundary}$ & 0.84 & 0.96 & 0.90 \\
    $Recall_{boundary}$ & 0.94 & 0.98 & 0.96 \\
    \hline
  \end{tabular}
  \label{table4}
\end{table}
\subsection{Comparison with Other Methods}

We compared our method with other approaches in two key aspects: (1) the accuracy of room semantic segmentation and (2) the precision and recall of reconstructed floorplans. These comparisons were conducted on both the GibLayout and ISPRS datasets. For room semantic segmentation, we compared our method with the Mask R-CNN model trained in the FloorSP method \cite{chen2019floor}. For floorplan reconstruction, we compared our approach with both the FloorSP, Heat\cite{chen2022heat}, RoomFormer\cite{yue2023connecting}, PolyGraph\cite{sun2025polygraph}, and PolyRoom\cite{liu2024polyroom} methods.
\par
\begin{figure*}
    \centering
    \includegraphics[width=0.8\textwidth]{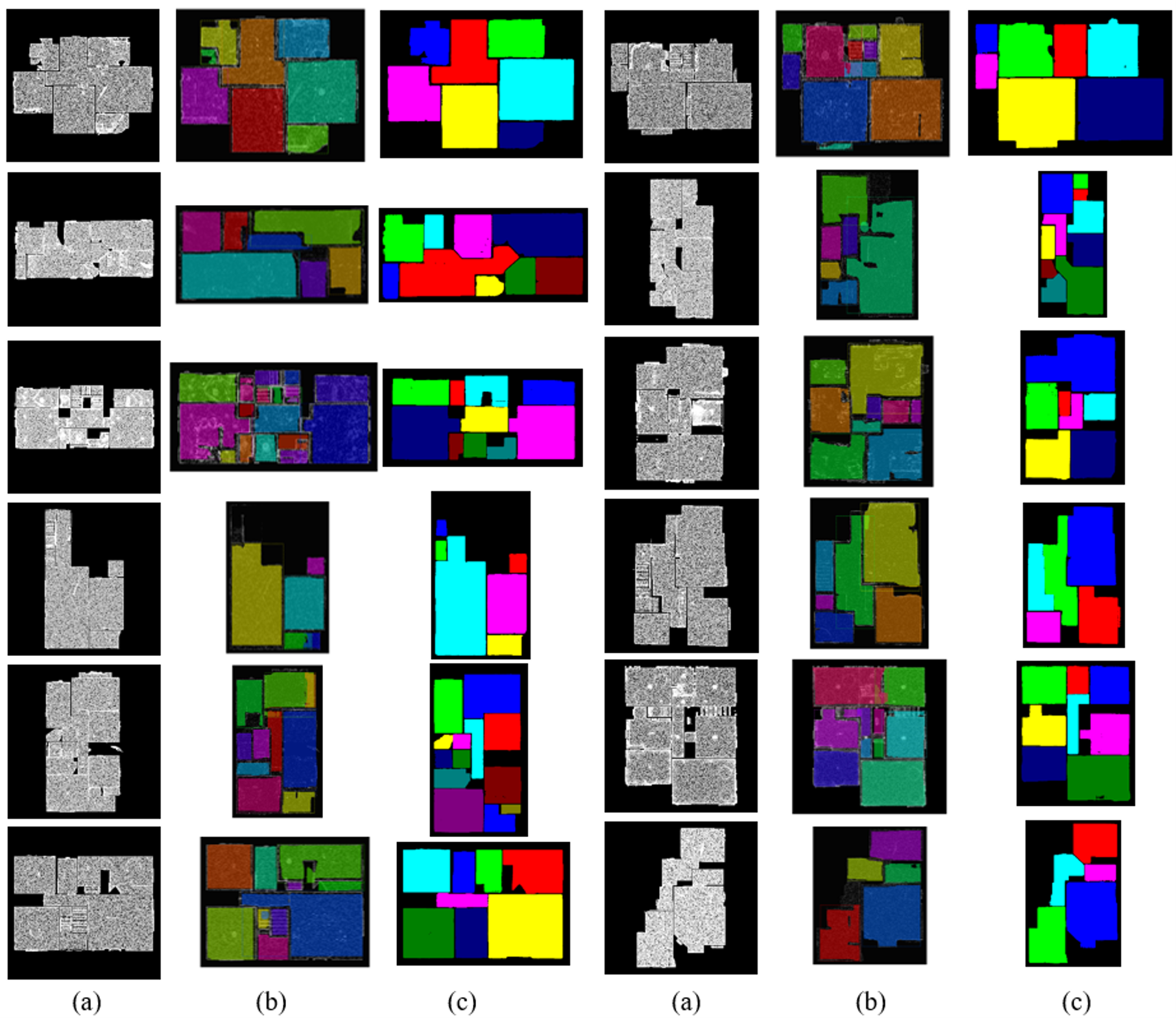}
    \caption{The room segmentation effect of the Giblayout dataset. (a) represents the input density map; (b) represents the result of the room segmentation of the Mask R-CNN model trained by floorsp; (c) represents the segmentation effect of our method.}
    \label{experience7}
\end{figure*}

\begin{figure}
    \centering
    \includegraphics[width=0.9\linewidth]{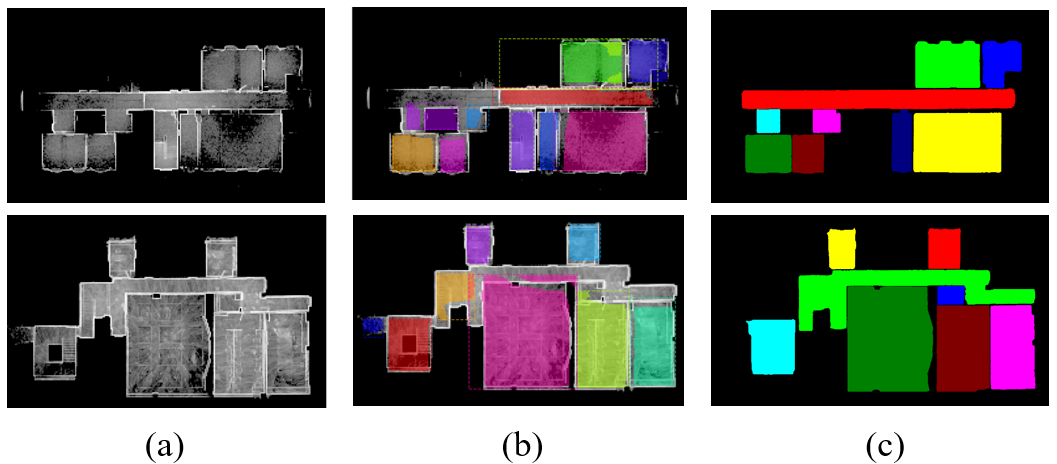}
    \caption{The room segmentation effect of the ISPRS dataset. (a) shows the input density map; (b) shows the result of the room segmentation by the Mask R-CNN model trained by floorsp; (c) shows the segmentation effect of our method.}
    \label{experience8}
\end{figure}

 First, we evaluated our SAM-based room segmentation method against the Mask R-CNN model from FloorSP. Figure~\ref{experience7} and Figure~\ref{experience8} present the room semantic segmentation results of the FloorSP Mask R-CNN pretrained model and our method on samples from the GibLayout and ISPRS datasets, respectively. These figures demonstrate that our method produces more complete and accurate masks. 
\par
\begin{figure*}
    \centering
    \includegraphics[width=0.9\textwidth]{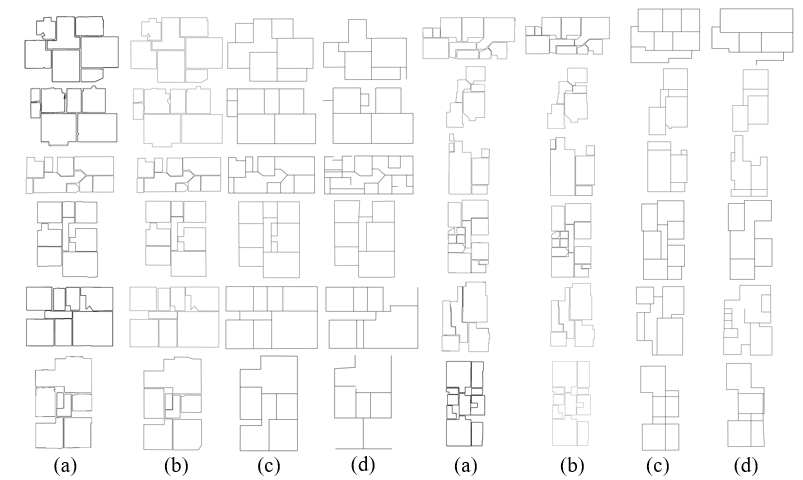}
    \caption{Comparison of floor plan reconstruction results of the Giblayout dataset. (a) represents the manually restored floor plan; (b) represents the result of our method; (c) represents the contour restored by the floorsp method; (d) represents the floor plan restored by the heat method.}
    \label{experience9}
\end{figure*}

\begin{figure}
    \centering
    \includegraphics[width=0.9\linewidth]{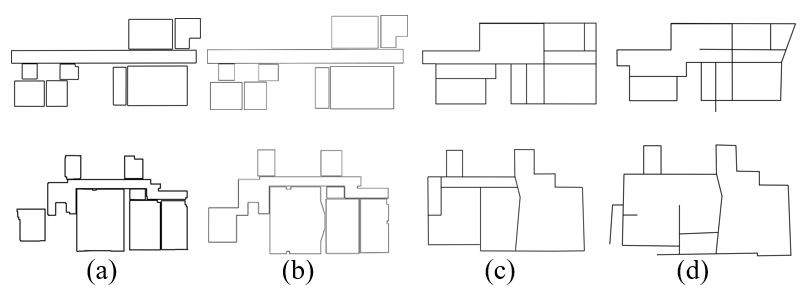}
    \caption{Comparison of floor plan reconstruction results for the ISPRS dataset. (a) shows the manually restored floor plan; (b) shows the result of our method; (c) shows the contours restored by the floorsp method; (d) shows the floor plan restored by the heat method.}
    \label{experience10}
\end{figure}

 Next, we compared floorplan reconstruction performance with FloorSP and Heat. Figure~\ref{experience9} and Figure~\ref{experience10} show the reconstructed floorplans generated by FloorSP, Heat, and our method on the GibLayout and ISPRS benchmark datasets, respectively, alongside the manually drawn ground truth floorplans. Our method produces floorplans with higher completeness and richer details.
To evaluate the effectiveness of room floor plan reconstruction, we compare the recall and precision of the reconstructed room floor plans of FloorSP, Heat, RoomFormer, PolyGraph, and PolyRoom methods. As shown in Table~\ref{table5}, our method achieves higher room floor plan recall and precision on both datasets, indicating that our method reconstructs more complete and accurate rooms compared to other methods.


\begin{figure}
    \centering
    \includegraphics[width=0.8\linewidth]{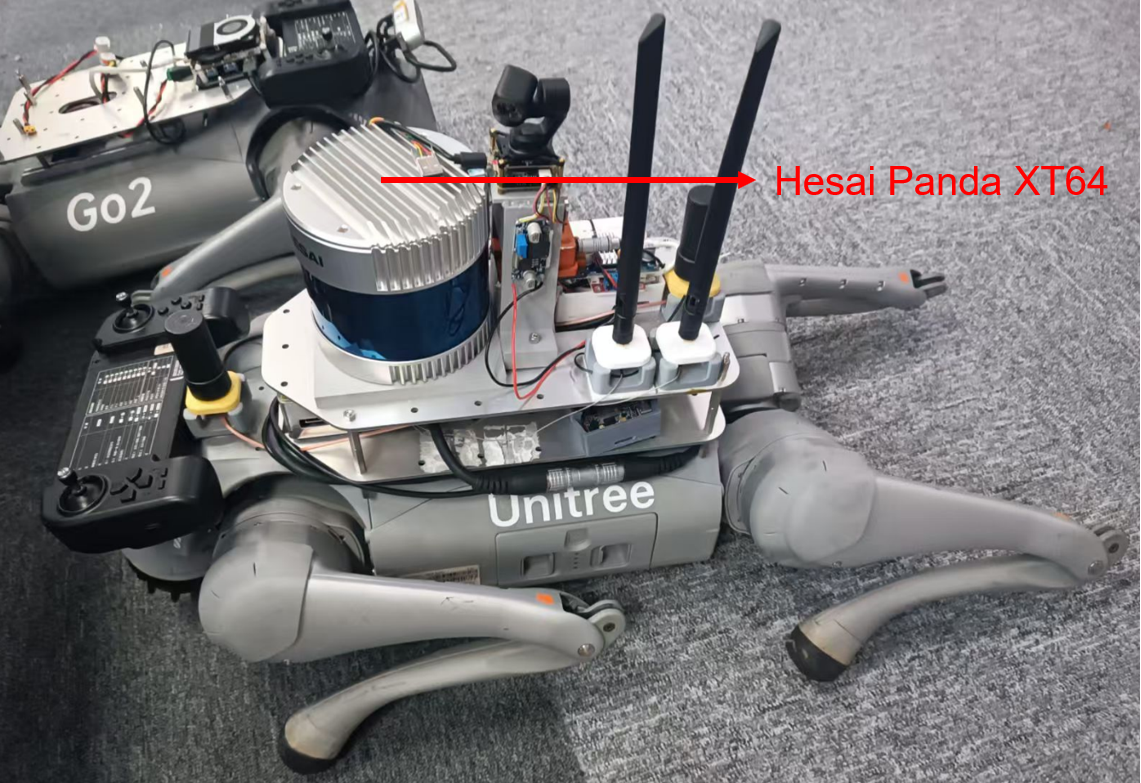}
    \caption{  Experimental platform, we built a robot dog platform, installed the sensors required for mapping, and equipped it with a self-developed mapping algorithm.}
    \label{experience14}
\end{figure}
\begin{figure}
    \centering
    \includegraphics[width=1\linewidth]{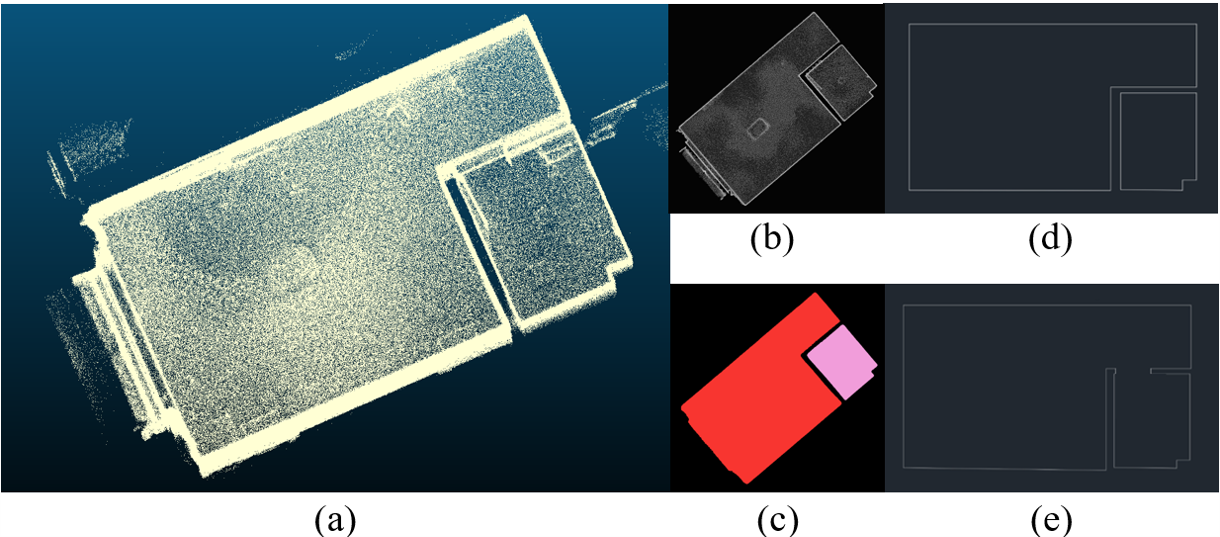}
    \caption{ Real data experiment 1. (a) represents the input point cloud; (b) represents the point cloud density map; (c) represents the segmentation results of each room; (d) represents the drawn outline; (e) represents the plane map after the topological relationship is restored.}
    \label{experience12}
\end{figure}

\begin{figure}
    \centering
    \includegraphics[width=1\linewidth]{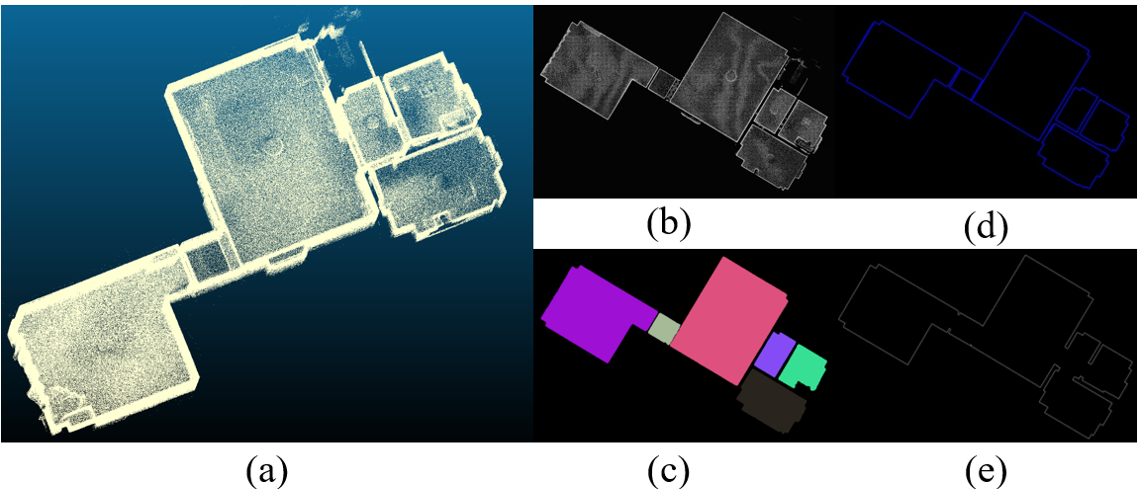}
    \caption{  Real data experiment 2. (a) represents the input point cloud; (b) represents the point cloud density map; (c) represents the segmentation results of each room; (d) represents the drawn outline; (e) represents the plane map after the topological relationship is restored.}
    \label{experience13}
\end{figure}

\begin{table}
  \centering
  \caption{Comparison of recall and precision of our method with floorsp and heat methods in room floor plan reconstruction.}
  \renewcommand{\arraystretch}{1.5}
  \begin{tabular}{c@{\hspace{3em}}c@{\hspace{3em}}c@{\hspace{3em}}c@{\hspace{3em}}c}
    \toprule
    Datasets & \multicolumn{2}{c}{Giblayout} & \multicolumn{2}{c}{ISPRS benchmark} \\
    \cmidrule(lr){2-3} \cmidrule(lr){4-5}
    Score & Precision & Recall & Precision & Recall \\
    Floor-sp & 0.74 & 0.77 & 0.88 & 0.86 \\
    Heat & 0.72 & 0.74 & 0.73 & 0.73 \\
    RoomFormer & 0.80 & 0.82 & 0.85 & 0.87 \\
    PolyGraph & 0.78 & 0.79 & 0.90 & 0.88 \\
    PolyRoom & 0.76 & 0.81 & 0.89 & 0.85 \\
    Ours & \textbf{0.92} & \textbf{0.94} & \textbf{0.94} & \textbf{0.95} \\
    \bottomrule
  \end{tabular}
  \label{table5}
\end{table}

\subsection{Real-World Experiments}
To further verify the effectiveness of our method, as shown in Figure~\ref{experience14}, we built a robot dog platform with the ability to build high-quality point cloud maps. We used it to collect multiple sets of indoor building data in real scenes and used our method to generate vector floor plan results. The most representative sets of data are shown in Figure~\ref{experience12} and Figure~\ref{experience13}.

\section{CONCLUSION}

This paper proposes a LiDAR point cloud-based indoor floorplan reconstruction method that integrates point cloud density maps with SAM's zero-shot segmentation to achieve efficient and robust reconstruction. Compared to traditional geometric approaches and Mask R-CNN-based deep learning methods, our approach improves performance through: 1) generating bird's-eye view density maps via grid-based filtering and adaptive resolution projection, combined with image enhancement techniques to mitigate noise and occlusions; 2) leveraging SAM's zero-shot segmentation alongside adaptive prompt points to produce high-quality single-room masks, thereby overcoming dependence on annotated data and enhancing generalization; and 3) employing mask-point cloud joint contour extraction and regularization to integrate semantic and geometric information, resulting in precise vector contours and accurate topological relationships. Experiments on the GibLayout and ISPRS datasets demonstrate the method's effectiveness across sparse point clouds, complex layouts, and large-scale scenes, achieving average precision and recall above 0.90. Our approach outperforms traditional methods, particularly in non-Manhattan structures and noisy environments. However, the method still faces challenges with point clouds lacking ceiling data. Future work will incorporate ground semantics to improve segmentation generalization and utilize image-based segmentation to recover window structures.

\printbibliography
\end{document}